\documentclass[USenglish,twocolumn]{article}
\usepackage[utf8]{inputenc}
\usepackage[big,online]{dgruyter}
\usepackage[sort&compress,square,numbers]{natbib}
\usepackage{times}

\begin{document}

%%%--------------------------------------------%%%
%%% Please do not alter the following lines: %%%
%%%--------------------------------------------%%%
	%\articletype{Proceedings}
  %\aop
  \DOI{10.1515/}
  \openaccess
  \pagenumbering{gobble}
%%%--------------------------------------------%%%

\title{Tracking Any Point Methods for Markerless 3D Tissue Tracking in Endoscopic Stereo Images}
\runningtitle{Tracking Any Point for 3D Tracking}
%\subtitle{Insert subtitle if needed}

\author[1,2]{K. Reuter}
\author[1]{S. Guttikonda}
\author[1]{S. Latus}
\author[1]{L. Maack}
\author[2]{C. Betz}
\author[3,4]{T. Maurer}
\author[1]{A. Schlaefer} 
\runningauthor{K.Reuter et al.}

\affil[1]{\protect\raggedright Institute of Medical Technology and Intelligent Systems, Hamburg University of Technology, Hamburg, Germany, e-mail: konrad.reuter@tuhh.de}
\affil[2]{\protect\raggedright Department of Otorhinolaryngology, Head and Neck Surgery and Oncology, University Medical Center Hamburg-Eppendorf, Hamburg, Germany}
\affil[3]{\protect\raggedright Martini-Klinik Prostate Cancer Center, University Medical Center Hamburg-Eppendorf, Hamburg, Germany}
\affil[4]{\protect\raggedright Department of Urology, University Medical Center Hamburg-Eppendorf, Hamburg, Germany}

\abstract{Minimally invasive surgery presents challenges such as dynamic tissue motion and a limited field of view. Accurate tissue tracking has the potential to support surgical guidance, improve safety by helping avoid damage to sensitive structures, and enable context-aware robotic assistance during complex procedures. In this work, we propose a novel method for markerless 3D tissue tracking by leveraging 2D Tracking Any Point (TAP) networks. Our method combines two CoTracker models, one for temporal tracking and one for stereo matching, to estimate 3D motion from stereo endoscopic images. We evaluate the system using a clinical laparoscopic setup and a robotic arm simulating tissue motion, with experiments conducted on a synthetic 3D-printed phantom and a chicken tissue phantom. Tracking on the chicken tissue phantom yielded more reliable results, with Euclidean distance errors as low as 1.1 mm at a velocity of 10 mm/s. These findings highlight the potential of TAP-based models for accurate, markerless 3D tracking in challenging surgical scenarios.}

\keywords{CoTracker, Tracking Any Point, Stereo Vision, Endoscopic Surgery, Minimally Invasive Surgery}

\maketitle

\section{Introduction}

Robot-assisted minimally invasive surgery allows for precise control of instruments through small incisions, reducing the risk of postoperative complication, blood loss, and mortality compared to open surgery \citep{davinci_advantages}. Patients also benefit from faster and more comfortable recoveries \citep{RAMIS_advantages}. As a result, systems like the da Vinci Surgical System by Intuitive Surgical are seeing growing adoption, with a 17\% increase in procedures reported in 2024 \cite{davinci_report}.

Despite these advantages, navigation and visual tracking remain challenging due to constant tissue motion caused by breathing, heartbeat, and tool interaction, combined with a limited endoscopic field of view. Real-time, automatic tissue tracking could help mitigate these challenges, for example by continuously localizing target tissue or providing warnings near sensitive structures, thus enhancing surgical precision and safety. While external tracking systems offer high accuracy, marker placement is impractical in minimally invasive settings, making markerless approaches using only endoscopic images highly promising.

Recent image-based tracking methods increasingly rely on deep learning. In particular, Tracking Any Point (TAP) models like CoTracker3 \citep{cotracker3} have demonstrated strong performance in tracking arbitrary 2D points across video sequences. However, these methods lack the 3D spatial understanding necessary for surgical applications.

Stereo vision offers a practical approach to 3D reconstruction by estimating depth through disparity computation. Classical stereo matching techniques often struggle on texture-less surfaces and rely on local features. Deep learning-based approaches improve robustness by incorporating global context but tend to be limited by their training domains \citep{DL_limits}.

In this work, we propose a novel stereo matching approach that leverages TAP networks for markerless 3D tissue tracking. Our method aims to combine the domain generality of classical techniques with the accuracy and robustness of learning-based models. We evaluate our system using a clinical laparoscopic setup and a robotic arm to simulate tissue motion, with experiments conducted on both a 3D-printed and a chicken tissue phantom.

\section{Methods}

\subsection{Proposed Method}

\begin{figure}
    \centering
    \includegraphics[width=\linewidth]{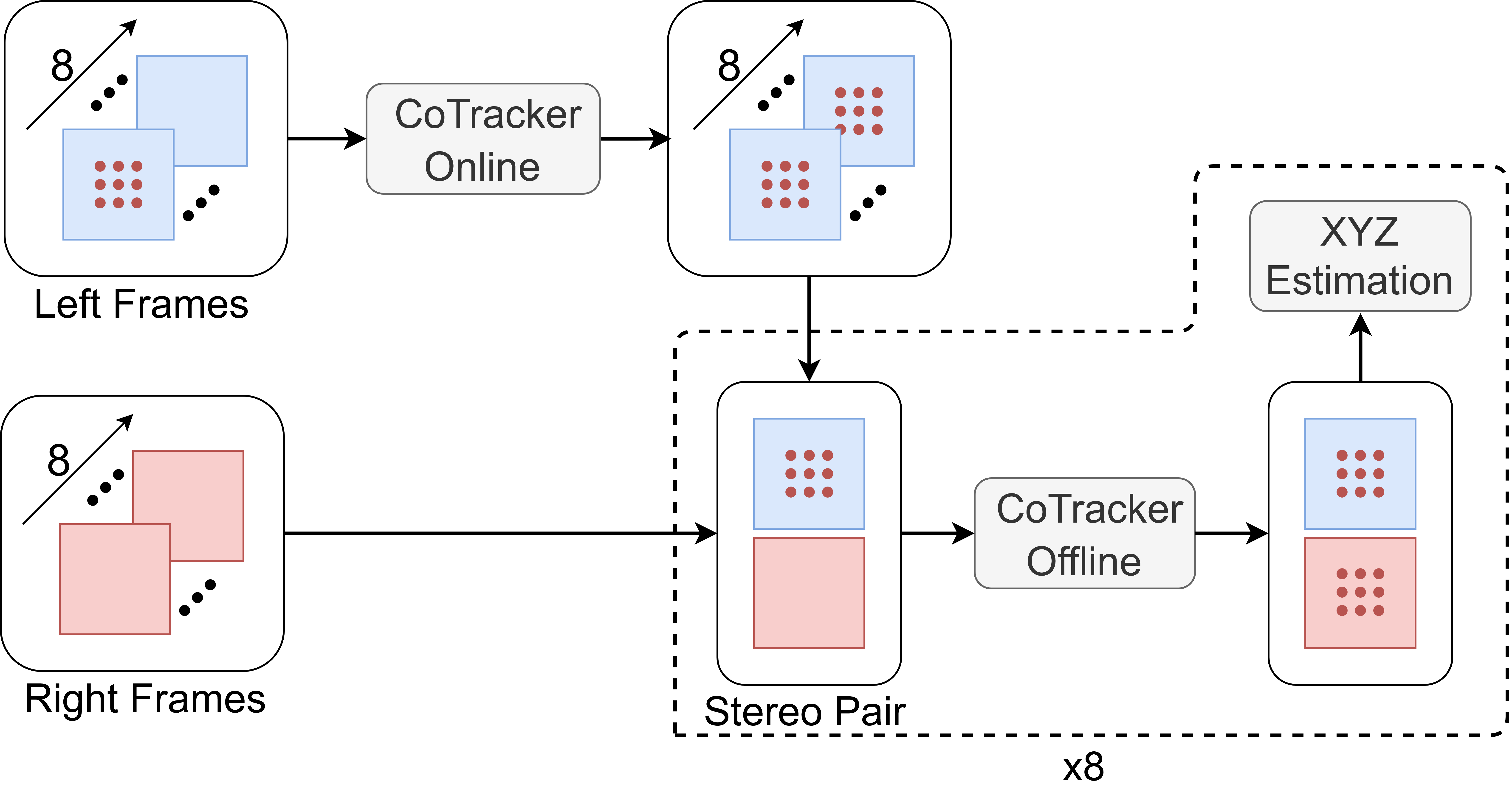}
    \caption{Overview of the proposed method. Eight frames from the left and right camera are received as the input. The tracking points are initialized in the first frame of the left camera. A first CoTracker model is used to track the points along the temporal dimension. Afterwards the frames are stacked as stereo pairs. A second CoTracker model is used to perform stereo matching and find the tracking points in the frames of the right camera. The disparity between points in the left and right frame can then be used to calculate the depth (Z) and subsequently the lateral position (X, Y).}
    \label{fig:method}
\end{figure}

Stereo matching identifies corresponding pixels between two views of the same scene. TAP methods, originally developed for temporal point tracking, perform a similar correspondence task and can be adapted for spatial matching. We repurpose TAP for stereo vision by replacing the temporal video input with a stereo image pair. By combining two CoTracker3 models, one for temporal tracking and one for stereo correspondence, we propose a novel 3D tracking method based entirely on 2D tracking algorithms. An overview of our method is shown in Figure \ref{fig:method}. Given a stream of rectified stereo image pairs, the tracking target is defined in the first frame of either camera. CoTracker3 comes in two versions: an online version, processing batches of eight frames, and an offline version, processing all frames at once. To track the target over time in the same view, we use the online CoTracker3 model. Stereo correspondence is handled by the offline CoTracker3 model, applied to each pair to find matching points in the opposing view. From these spatial correspondences, we estimate the depth (along the Z-axis) as well as the lateral position on the X and Y axis. Instead of tracking a single points, we use a small ensemble of points to mitigate depth estimation errors. Along all axis we derive the single position estimate by computing the median over all points. Finally, the values are converted from camera units into mm.

\subsection{Experimental Setup}

\begin{figure}
    \centering
    \includegraphics[width=.8\linewidth]{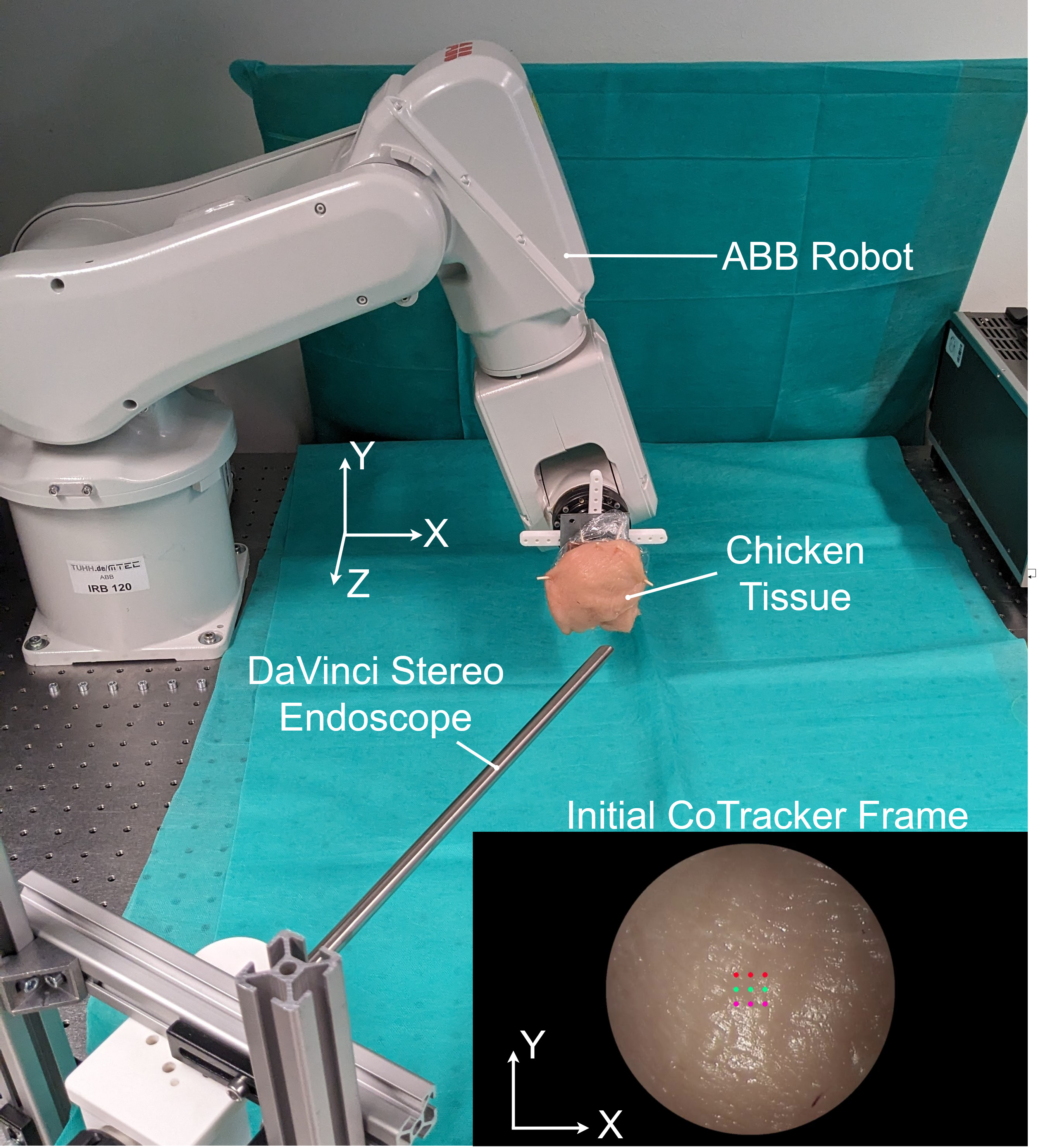}
    \caption{The setup used for tracking evaluation, consisting of a DaVinci Stereoscope and a robotic arm to hold and move tissue samples. In the bottom right corner the view from one of the cameras is shown.}
    \label{fig:setup}
\end{figure}

For image acquisition, we use a da Vinci laparoscope equipped with stereo cameras (1080 $\times$ 1920 pixels, 30 FPS). The laparoscope tip is placed approximately 100 mm from the target, mimicking clinical working distances \citep{lapro_distance}, and yielding a field of view of 60 $\times$ 105 mm. A robotic arm (ABB IRB-120) simulates tissue motion by moving the target along predefined trajectories while logging ground-truth positions. The setup is illustrated in Figure \ref{fig:setup}. To align the camera and robot coordinate systems, a checkerboard is mounted on the end effector and moved to 125 known positions. We estimate the position of the checkerboard in each image to calculate an affine transformation from camera to robot coordinate system.

\begin{figure}
\centering
\begin{minipage}[t]{0.375\linewidth}
    \includegraphics[width=\textwidth]{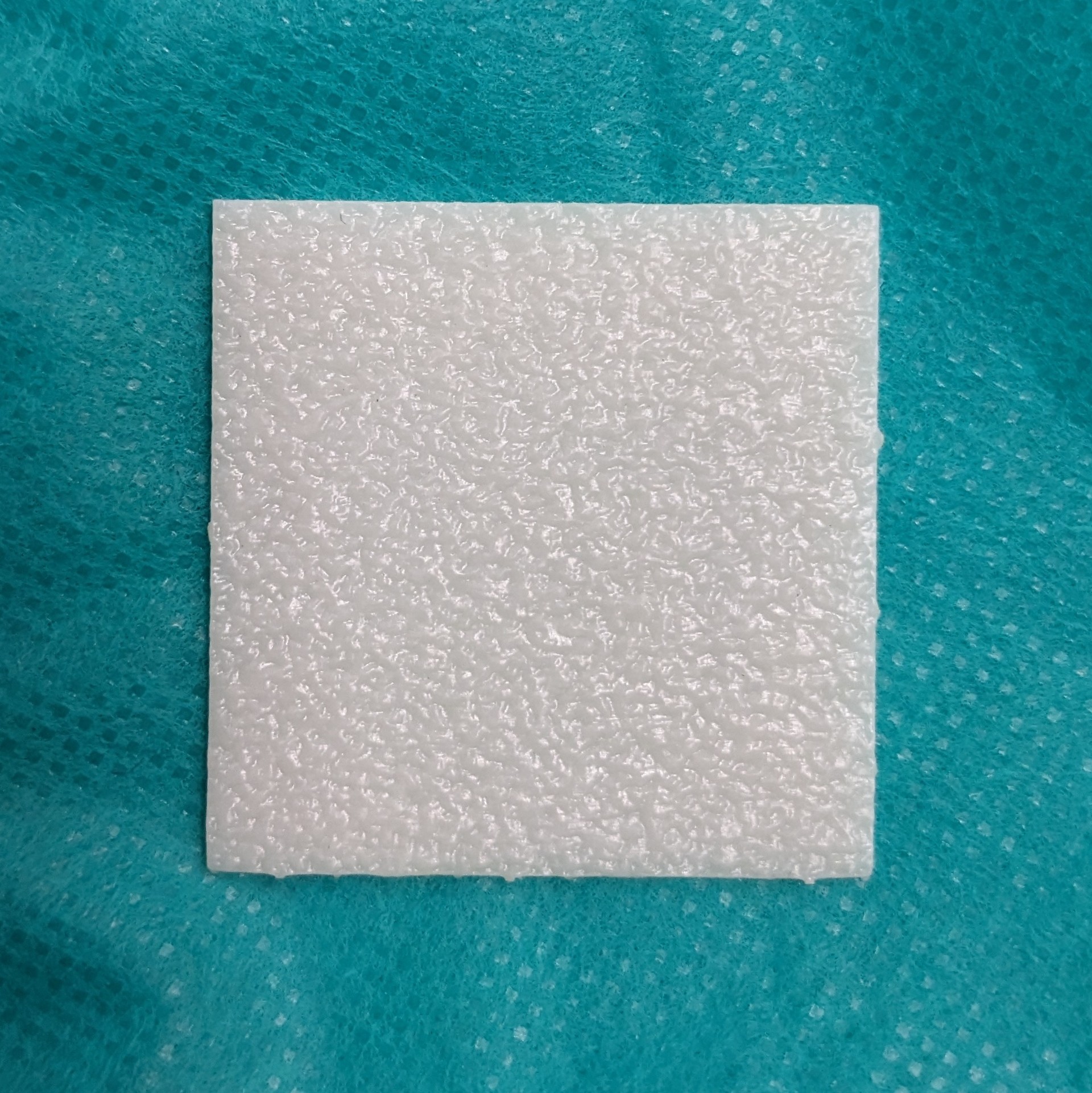}
\end{minipage}\begin{minipage}[t]{0.375\linewidth}
    \includegraphics[width=\textwidth]{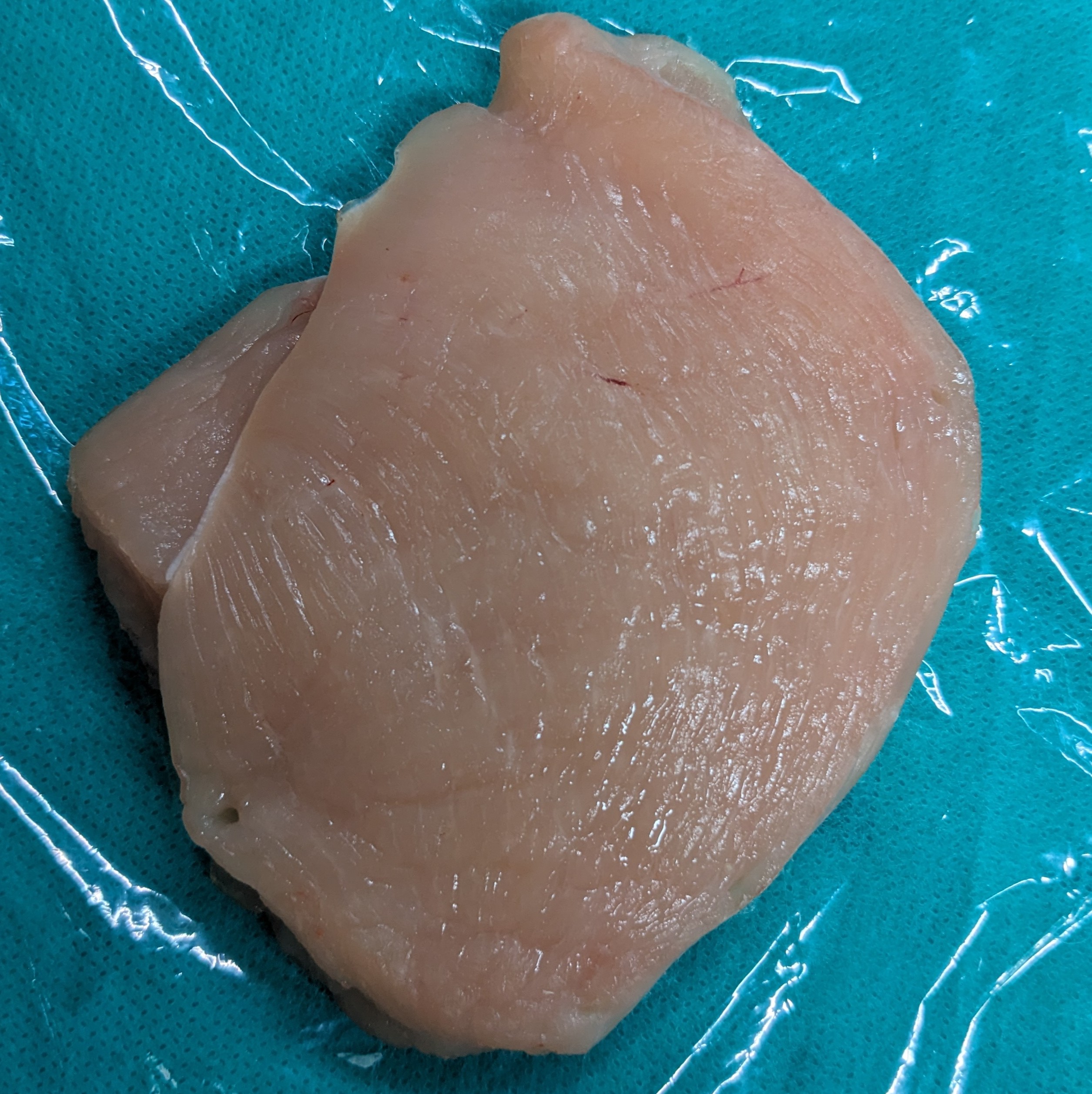}
\end{minipage}
\caption{Tracking Targets. Left: 3D print with randomly structured surface. Right: Chicken breast.}
\label{fig:phantoms}
\end{figure}

We evaluate the system's performance using a 3D-printed phantom and a chicken tissue phantom. The 3D printed phantom has a structured surface but no distinct features, simulating a specifically challenging tracking target. The chicken tissue is chosen to simulate real tissue. Both phantoms are shown in Figure \ref{fig:phantoms}. For each run, we select a 100 $\times$ 100 pixel template from the center of the first frame as the tracking target. A 3 $\times$ 3 grid of points is placed over this target area. The robot's end effector follows a randomized trajectory over a 30-second period, constrained within a 30 $\times$ 30 $\times$ 30 mm volume.

The evaluation consists of three stages: (1) Z-axis (depth) motion using the second CoTracker for stereo matching; (2) XY-plane motion using the first CoTracker for temporal tracking; and (3) full 3D motion tracking using the complete framework. The velocities range from 10 mm/s to 80 mm/s, with three unique trajectories per velocity. Additionally, we assess how query point density affects accuracy by varying the grid size from a single point up to 10 $\times$ 10, focusing on chicken tissue at 50 mm/s.

We subtract the initial positions from the tracked and robot paths for evaluation, focusing on relative movement. The primary metric is the mean Euclidean distance between the tracked and the ground-truth robot path, calculated along the axis where movement was introduced. To simulate occluded target boundaries, as it often occurs in real surgery, we overlay a virtual mask on the images. A sample of the initial frame is shown in Figure \ref{fig:setup}.

\section{Results}

The transformation between the camera and robot coordinate systems yields a residual Euclidean error of 0.124 ± 0.062 mm when estimating previously unseen checkerboard positions. On an Nvidia RTX 4090 GPU, our tracking framework achieves real-time performance, processing 33 frames per second (FPS) with a 3 $\times$ 3 grid of points.

\begin{figure}
    \centering
    \includegraphics[width=\linewidth]{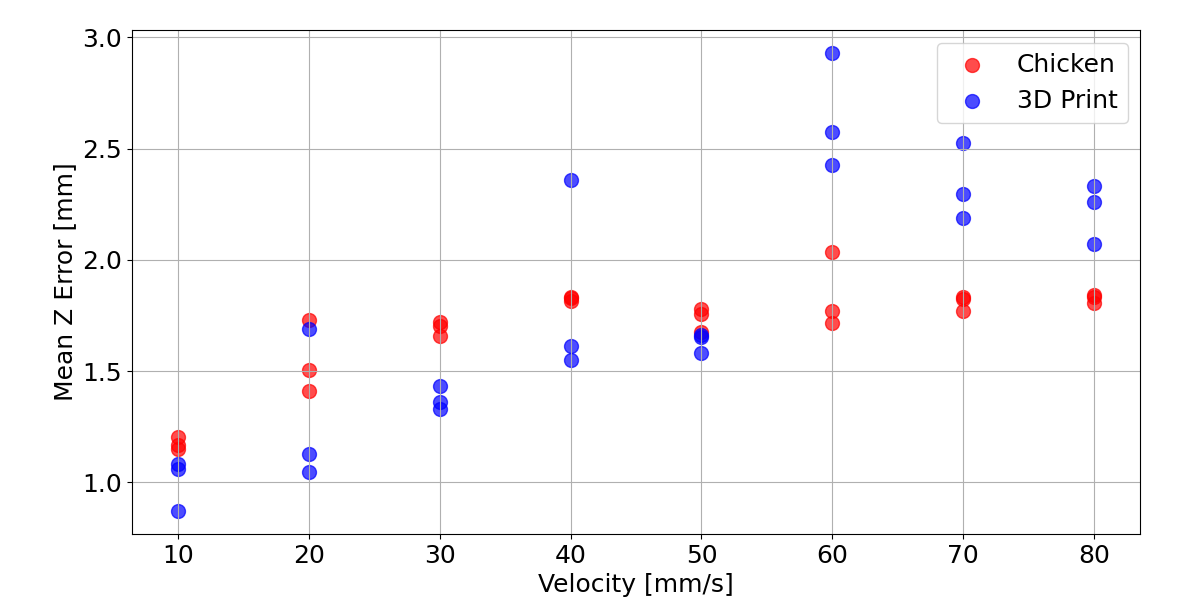}
    \caption{Mean error for movements along the Z (depth) axis in mm across different velocities.}
    \label{fig:Z}
\end{figure}

Figure~\ref{fig:Z} shows the results for Z-axis (depth) motion. For the 3D-printed phantom, depth errors start around 1 mm at low velocities but increase steadily, surpassing 2 mm for speeds above 50 mm/s. In contrast, the chicken tissue phantom exhibits slightly higher errors at lower speeds but maintains depth accuracy below 2 mm even at 80 mm/s.

\begin{figure}
    \centering
    \includegraphics[width=\linewidth]{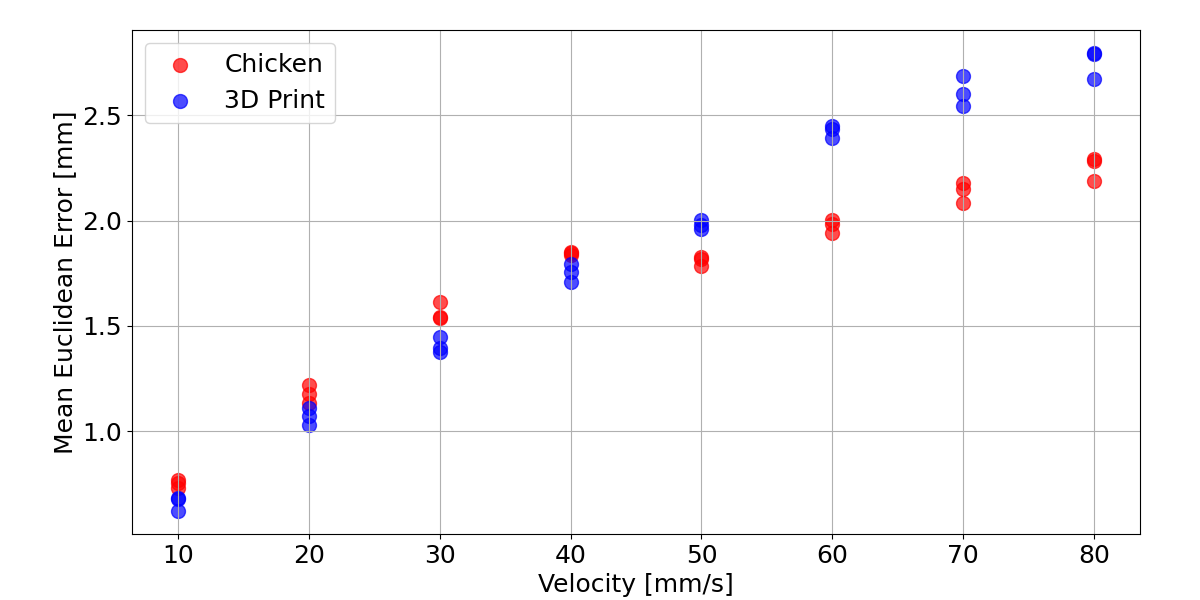}
    \caption{Mean euclidean distance error in mm for movements along the X and Y axis (lateral dimensions) across different velocities.}
    \label{fig:XY}
\end{figure}

The XY-plane tracking results in Figure~\ref{fig:XY} demonstrate comparable accuracy between phantoms at low velocities, but errors increase more rapidly for the 3D-printed phantom with speed. It is noteworthy that there are no outliers. Across both phantoms, lateral tracking errors range from under 1 mm at 10 mm/s to over 2.5 mm at 80 mm/s.

\begin{figure}
    \centering
    \includegraphics[width=\linewidth]{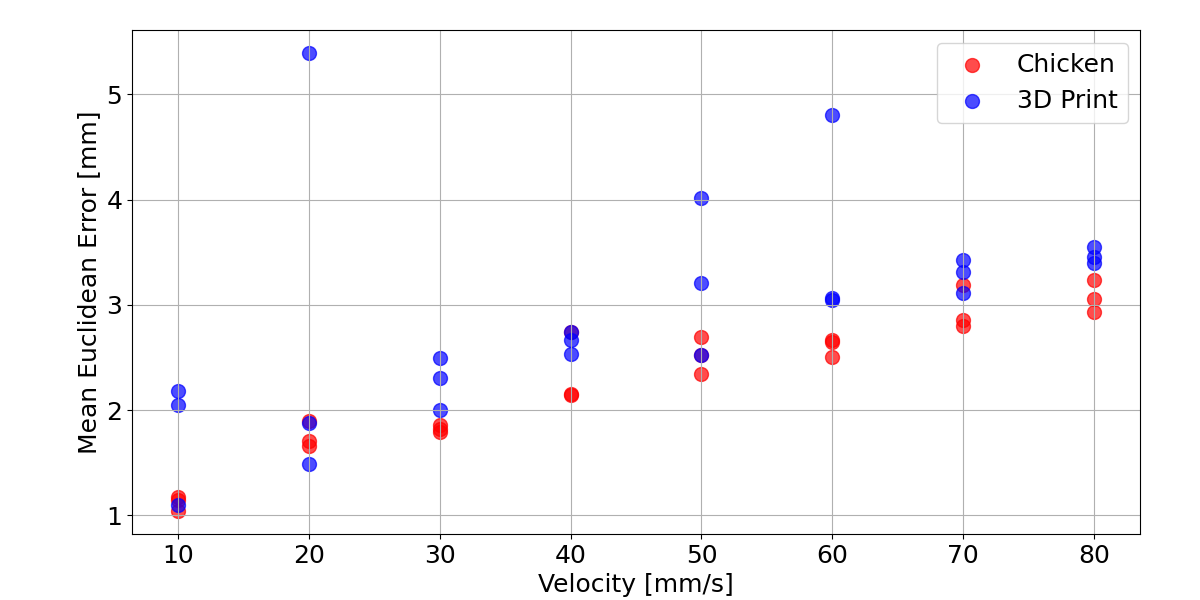}
    \caption{Mean euclidean distance error in mm for 3D movements across different velocities.}
    \label{fig:XYZ}
\end{figure}

For full 3D tracking, illustrated in Figure \ref{fig:XYZ}, the accumulated errors from Z and XY tracking become evident. Overall, tracking on the chicken breast phantom proves more reliable, with notably fewer outliers. In contrast, the 3D-printed phantom exhibits occasional large errors exceeding 4 mm. Ignoring these outliers, errors range from about 1.1mm for 10 mm/s to around 3.5mm at 80 mm/s.

\begin{figure}
    \centering
    \includegraphics[width=\linewidth]{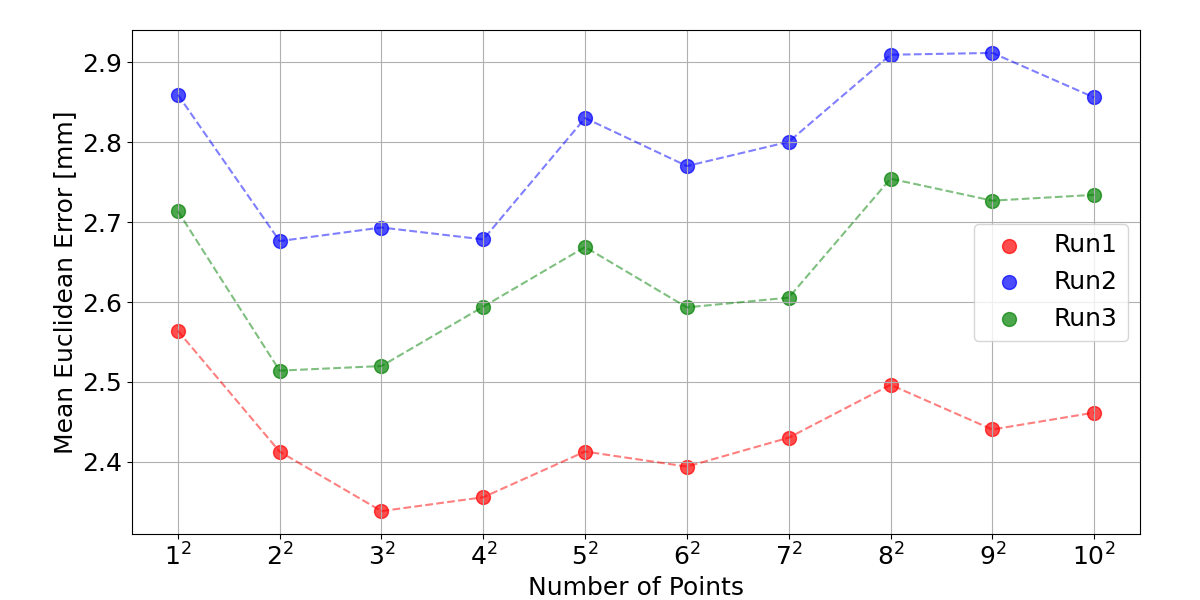}
    \caption{Mean euclidean distance error in mm for the chicken breast target at 50 mm/s across different numbers of tracking points.}
    \label{fig:Q}
\end{figure}

Figure \ref{fig:Q} analyzes the effect of query point density on accuracy. Initially, increasing the number of points reduces the tracking error, with optimal results achieved around 9 query points. Beyond this point, performance degrades. In runs 2 and 3, configurations with more than 64 points even surpass the error of the single-point baseline. We also observe a slight decrease in frame rate with increasing point density. For instance, using a 10 $\times$ 10 grid of tracking points results in approximately 29 FPS.

\section{Discussion}

Our results demonstrate the general feasibility of using 2D TAP methods for stereo matching and extending them into 3D tracking. Overall, tracking on chicken tissue proves more stable than on the 3D-printed phantom. We consider multiple reasons for this. First, the chicken tissue exhibits more distinct features, making it easier for CoTracker to lock onto consistent visual landmarks. In contrast, the 3D print surface is relatively feature-poor, which becomes especially problematic at higher velocities, where motion blur further reduces feature distinctiveness. Additionally, the white plastic surface reflects light strongly, introducing specular reflections that often interfere with tracking.

Despite these challenges, we did not observe complete tracking failures. The temporal CoTracker performed robustly across all tested velocities and handled partial occlusions from reflections well. While the stereo CoTracker occasionally produced inaccurate matches, these led to localized tracking errors rather than a full pipeline breakdown.

Beyond robustness, our method offers several advantages. Thanks to the generality of the CoTracker architecture, it can be applied to diverse computer vision tasks, provided the stereo setup is calibrated. As TAP methods evolve, improvements can directly benefit our pipeline without architectural changes. Although we focus on square tracking regions in this work, the method is theoretically capable of handling arbitrarily shaped objects. 

Nonetheless, our method has limitations. The current tracking errors are still relatively high for clinical application. Tracking points are currently uniformly sampled within the region of interest. Selecting more informative or discriminative points could further improve performance. Moreover, our results indicate a trade-off between the number of tracking points and accuracy, with the optimal configuration in our experiments occurring around nine points. Arbitrarily increasing point density does not yield better results and may even introduce noise or redundancy. 
Additionally, TAP methods like CoTracker are not inherently designed for stereo matching, which involves only two images, possibly leading to suboptimal results. As our method uses two CoTracker instances, each image from one camera stream is currently processed twice, resulting in increased computational load. Future work should explore using a shared encoder to reduce redundancy and improve efficiency. Finally, while our results are promising on controlled laboratory data, further validation on real surgical scenes is needed to assess practical applicability.

\section{Conclusion}

We present a method for 3D tracking by combining two successive 2D TAP models, one for temporal tracking and one for stereo correspondence. The approach shows robust performance across different velocities. Future work will focus on optimizing point selection, reducing computational requirements, and validating the method in real surgical environments.

\begin{acknowledgement}
We would like to acknowledge Intuitive Surgical for providing the equipment needed for our experiments. 
\end{acknowledgement}

\textsf{\textbf{Author Statement}}

This research was co-funded by the MARLOC project (DFG, grant SCHL 1844-10-1) and by the European Union under Horizon Europe programme grant agreement No. 101059903; and by the European Union funds for the period 2021-2027. Tobias Maurer: Research funding from Brainlab, Intuitive Surgical, and Telix. Conflict of interest: Authors state no conflict of interest. Informed consent: Informed consent has been obtained from all individuals included in this study. Ethical approval: not applicable. 

%\bibliographystyle{...}
%\bibliography{...}

\end{document}